\documentclass{article}
\pdfoutput=1

\PassOptionsToPackage{numbers, square}{natbib}
\usepackage[preprint]{style}

\usepackage[utf8]{inputenc} 
\usepackage[T1]{fontenc}    
\usepackage{hyperref}       
\usepackage{url}            
\usepackage{booktabs}       
\usepackage{amsfonts}       
\usepackage{nicefrac}       
\usepackage{microtype}      
\usepackage{xcolor}         
\usepackage{multirow}
\usepackage{natbib}
\usepackage{bm}
\usepackage{amssymb} 
\usepackage[pdftex]{graphicx}
\usepackage{amsmath,amsthm,amssymb,amsfonts}
\usepackage{mathtools}
\usepackage{floatrow}
\usepackage{subfig}
\newfloatcommand{capbtabbox}{table}[][\FBwidth]
\title{STD-PLM: Understanding Both Spatial and Temporal Properties of Spatial-Temporal Data with PLM}

\author{
\textbf{Yiheng Huang}$^*$ \quad \textbf{Xiaowei Mao}$^*$ \quad  \textbf{Shengnan Guo} \quad \textbf{Yubin Chen} \\ \quad \textbf{Junfeng Shen} \quad \textbf{Tiankuo Li} \quad \textbf{Youfang Lin} \quad \textbf{Huaiyu Wan}$^\dagger$\\
\texttt{\{huangyiheng,maoxiaowei,guoshn,chenyubin,tanklee,jfshen,yflin,hywan\}@bjtu.edu.cn }\\
}

\begin{document}

\maketitle

\let\thefootnote\relax\footnote{$^*$These authors contribute equally to this work. $\dagger$ Corresponding authors.}

\begin{abstract}
Spatial-temporal forecasting and imputation are important for real-world intelligent systems. Most existing methods are tailored for individual forecasting or imputation tasks but are not designed for both. Additionally, they are less effective for zero-shot and few-shot learning. While pre-trained language model (PLM) have exhibited strong pattern recognition and reasoning abilities across various tasks, including few-shot and zero-shot learning, their applications in spatial-temporal data understanding has been constrained by insufficient modeling of complex correlations such as the temporal correlations, spatial connectivity, non-pairwise and high-order spatial-temporal correlations within data. In this paper, we propose STD-PLM for understanding both spatial and temporal properties of \underline{S}patial-\underline{T}emporal \underline{D}ata with \underline{PLM}, which is capable of implementing both spatial-temporal forecasting and imputation tasks. STD-PLM understands spatial-temporal correlations via explicitly designed spatial and temporal tokenizers. Topology-aware node embeddings are designed for PLM to comprehend and exploit the topology structure of data in inductive manner. Furthermore, to mitigate the efficiency issues introduced by the PLM, we design a sandglass attention module (SGA) combined with a specific constrained loss function, which significantly improves the model's efficiency while ensuring performance. Extensive experiments demonstrate that STD-PLM exhibits competitive performance and generalization capabilities across the forecasting and imputation tasks on various datasets. Moreover, STD-PLM achieves promising results on both few-shot and zero-shot tasks. The code is made available at \href{https://github.com/Hyheng/STD-PLM}{https://github.com/Hyheng/STD-PLM}
\end{abstract}

%

\section{Introduction}
Understanding both spatial and temporal properties of spatial-temporal data is crucial for various real-world dynamic systems such as intelligent transportation ~\cite{traffictheory} and urban planning ~\cite{trafficplanning}. In practice, spatial-temporal forecasting and imputation are the two most pivotal and common tasks. Specifically, precise spatial-temporal forecasting aids in effective traffic management and travel planning, while spatial-temporal imputation enables precise analysis of spatial-temporal patterns and supporting other dependent tasks. Although extensive studies have achieved satisfactory accuracy in spatial-temporal forecasting and imputation, they rely on extensive historical data for training. However, obtaining comprehensive datasets for all the studied regions is challenging due to the high cost of collecting long-term data. Consequently, the zero-shot ~\cite{zeroshotlearning} and few-shot learning ~\cite{few-shotlearning} capabilities are hindered for spatial-temporal forecasting methods across different regions.
Moreover, existing methods are usually tailored to specific tasks and not designed for both forecasting and imputation. Each method requires domain expertise and task-specific designs, increasing costs and complicating the deployment of forecasting and imputation methods. To summarize, we need a method that possesses powerful zero-shot and few-shot learning capabilities and is versatile for both spatial-temporal forecasting and imputation tasks, making it practical and applicable in real-world scenarios.

We have noticed that PLM are widely recognized for strong performance in zero-shot and few-shot learning across a diverse range of tasks ~\cite{ge2024openagi}. However, due to the significant differences between spatial-temporal and textual data, it is challenging for PLM to comprehend spatial-temporal data. In addition, training a spatial-temporal PLM from scratch is non-trivial due to the limited availability of spatial-temporal data compared to textual data. To address this issue, researchers have attempted to adapt PLM for understanding spatial-temporal data ~\cite{STGLLM, STLLM}. However, these methods offer insufficient representation for spatial-temporal data. Thus, they are less effective for understanding both spatial and temporal properties of spatial-temporal data with PLM. First, existing PLM-based methods design tokens along the spatial dimension but ignore the tokens obtained from the temporal dimension. Second, topological connectivity information inherent in spatial-temporal data is less explored. Thirdly, existing methods do not take into account that under the large embedding dimensions of the PLM, as the number of tokens increases, the training and inference costs of the model will increase rapidly. On the other hand, existing PLM-based spatial-temporal forecasting methods primarily concentrate on forecasting tasks and overlook the task of imputation. This limitation restricts their versatility and efficiency for practical applications.

To solve the above limitations and challenges, we propose STD-PLM for understanding both spatial-temporal properties of \underline{S}patio-\underline{T}emporal \underline{D}ata with \underline{PLM}. First, we create spatial-temporal embeddings for fully capturing temporal correlations and exploiting the topology structure of data. Second, we develop a spatial tokenizer and a temporal tokenizer to activate the ability of pre-trained PLM to understand spatial-temporal data from both the spatial and temporal dimensions while incorporating the topology structure of data. Lastly, we design the sandglass attention module with a constrained loss function, which can effectively model more important and sparser non-pairwise and higher-order correlations, significantly reducing computational costs. Based on the explicitly design, STD-PLM is not only versatile for spatial-temporal forecasting and imputation but also exhibits accurate few-shot and zero-shot learning capabilities on spatial-temporal data. In summary, our contributions can be summarized as follows:

\begin{itemize}

    \item We propose STD-PLM for accurate spatial-temporal forecasting and imputation, as well as zero-shot and few-shot learning, by activating PLM to understand both the spatial and temporal properties of spatial-temporal data. 
    
    \item We develop spatial-temporal tokenizers to construct tokens from spatial and temporal dimensions while incorporating the topology structure, thereby enabling the PLM to perceive not only fine-grained spatial information but also coarse-grained overall temporal information. 
    
    \item We design the sandglass attention module and its constrained loss function, which significantly reduces computational overhead while ensuring performance. 
    
\end{itemize}

\section{Related Work}
\label{section:related}
\textbf{Spatial-Temporal Data Forecasting.} In recent years, a large number of models that can effectively model spatial-temporal dependencies have emerged. Several studies \cite{FC-LSTM} address temporal dynamics using RNNs. To deal with spatial dependency, the data can be divided into grids \cite{DeepSTN+, Periodic-CRN,ST-resnet, DMVST}, and then CNNs are utilized to capture spatial correlations. But not all data can be partitioned into grid form. In order to achieve a more general and effective model, researchers introduced the graph convolution model \cite{GCN,chebgcn} thereby implementing spatial feature aggregation based on adjacency matrices \cite{DCRNN, Stg2seq, STGCN, GRAPHWAVENET, STSGCN,AGCRN}. Models such as GMAN \cite{GMAN},ASTGCN \cite{ASTGCN} and ASTGNN \cite{ASTGNN} utilize attention mechanisms to further deal with temporal and spatial correlations. PDFormer \cite{PDFormer} proposed a delay-aware feature transformation module to explicit model the temporal delay of propagation. The current work has a fairly good forecasting accuracy, but most of the models do not take into account the few-shot learning, zero-shot learning.

\noindent \textbf{Spatial-Temporal Data Imputation.} Initially, in order to realize imputation in more complex scenes, successive works based on low-rank matrix completion \cite{mazumder2010spectral,yu2016temporal} were proposed. LATC \cite{LATC} effectively captures local and global trends in the data by combining low-rank matrix complementation with auto-regressive model. The development of deep learning has also contributed to the research on imputation. Brits \cite{BRITS} build an efficient self-supervised imputation model through bidirectional LSTM. E2GAN \cite{E2GAN}, GAIN \cite{GAIN}and other GAN \cite{GAN} based models transforms the imputation problem into a generative problem. In addition, some VAE \cite{VAE} based works \cite{mTAN} recovers missing values by modeling the distribution of the data in the hidden space. Recently the powerful effect of Diffusion Model \cite{DDPM} in the field of image generation attracts a large number of researchers. CSDI \cite{CSDI} is based on the Diffusion Model, which recovery of missing values from noise by a well-designed denoising network. Compared to CSDI, another diffusion model PriSTI \cite{PRISTI} enhances the utilization of conditional information and the construction of geographic prior knowledge. Existing imputation models face the same problem as forecasting models, both need to reduce the need for training samples and enhance generalization capabilities.

\noindent \textbf{Pre-trained Language Model.} PLM provide a powerful and unified framework for different tasks. There has been a significant amount of recent work validating the effectiveness of the PLM in other modalities. One Fits All(OFA) \cite{OFA} demonstrated the feasibility of using PLM for time series data through the experimental results of eight different temporal tasks, and put forward preliminary explanations. Articles such as Time-LLM \cite{TIMELLM,TEMPO} make more detailed modifications to PLM, using methods such as reprogramming, prompt pool, and contrast learning to align temporal and textual data. Models such as STLLM \cite{STLLM} and STGLLM \cite{STGLLM} enhance PLM's perception of spatial-temporal data by constructing tokens in the dimension of space. Existing PLM-based models have achieved certain results, confirming that PLM has the ability to handle spatial-temporal data. However, they underutilize spatial information and are difficult to accurately capture complex spatial dependencies, and there is much room for improvement.

\noindent \textbf{Sandglass Attention.} There are some prior works whose methods are similar to the sandglass attention, such as SSTBAN \cite{SSTBAN} and CrossFormer \cite{CrossFormer}, which both learn higher-order correlations through learnable reference points combined with an attention mechanism to  improve model efficiency. Although these studies have achieved good results, they still have some limitations, such as not fully utilizing the adjacency relationships in the original graph and being unable to achieve zero-shot learning between different graphs.


\section{Methodology}
\label{section:methodology}
\subsection{Problem Definition}
\noindent \textbf{Spatial-Temporal data.} Considering spatial-temporal data with \( T \) time slices and \( N \) nodes, we represent it as \( \bm{\mathcal{X}} = \{\bm{X}_{1}, \bm{X}_{2}, \ldots, \bm{X}_{T}\} \in \mathbb{R}^{T \times N \times C} \). Here, \( \bm{X}_{t} = \{ \bm{x}_{t,1}, \bm{x}_{t,2}, \ldots, \bm{x}_{t,N} \} \in \mathbb{R}^{N \times C} \), where \( \bm{x}_{t,n} \) represents the feature of node \( n \) at time \( t \), and \( C \) is the number of features. 

We use a binary mask tensor \( \bm{\mathcal{M}} \in \{0, 1\}^{T \times N \times C} \) to denote the positions of missing values in $\bm{\mathcal{X}}$, where \( m_{t, n, c} = 0 \) indicates that the data is missing, and \( m_{t, n, c} = 1 \) indicates that the data is observed.  To describe the relationships between nodes, we introduce a directed graph \( \mathcal{G} = ( \mathcal{V}, \mathcal{E}, \mathbf{A}) \). Here, \( V \) represents the set of nodes in the data. \( \mathcal{E} \) is the set of edges. \( \mathbf{A} \in \mathbb{R}^{N \times N} \) is the adjacency matrix.

\noindent \textbf{Forecasting task.} Given the historical spatial-temporal data with \( T \) time slices \(  \bm{\mathcal{X}} \) and the corresponding graph structure, the objective is to forecast future data \(  \{ \bm{X}_{t+1},\ldots, \bm{X}_{t+T-2}, \bm{X}_{t+T-1}\} \) for the next \( T \) time slices.\\


\noindent \textbf{Imputation task.} Given incomplete spatial-temporal data with \( T \) time slices \(  \bm{\mathcal{X}} \), mask tensor \( \bm{\mathcal{M}} \) and the corresponding graph structure, the goal is to estimate complete data \(  \{\bm{\hat X}_{t-T+1}, \ldots,\) \(\bm{\hat X}_{t-1},\) \(\bm{\hat X}_{t}\} \).\\

\noindent \textbf{Few-shot and Zero-shot task.} The term ”few-shot" refers to training a model based on a limited amount of data. The term "zero-shot" denotes the direct testing of a model trained on a source dataset on a destination dataset, without any additional training.

\subsection{Model Structure}
\begin{figure*}[h]
  \centering
  \includegraphics[width=1\textwidth]{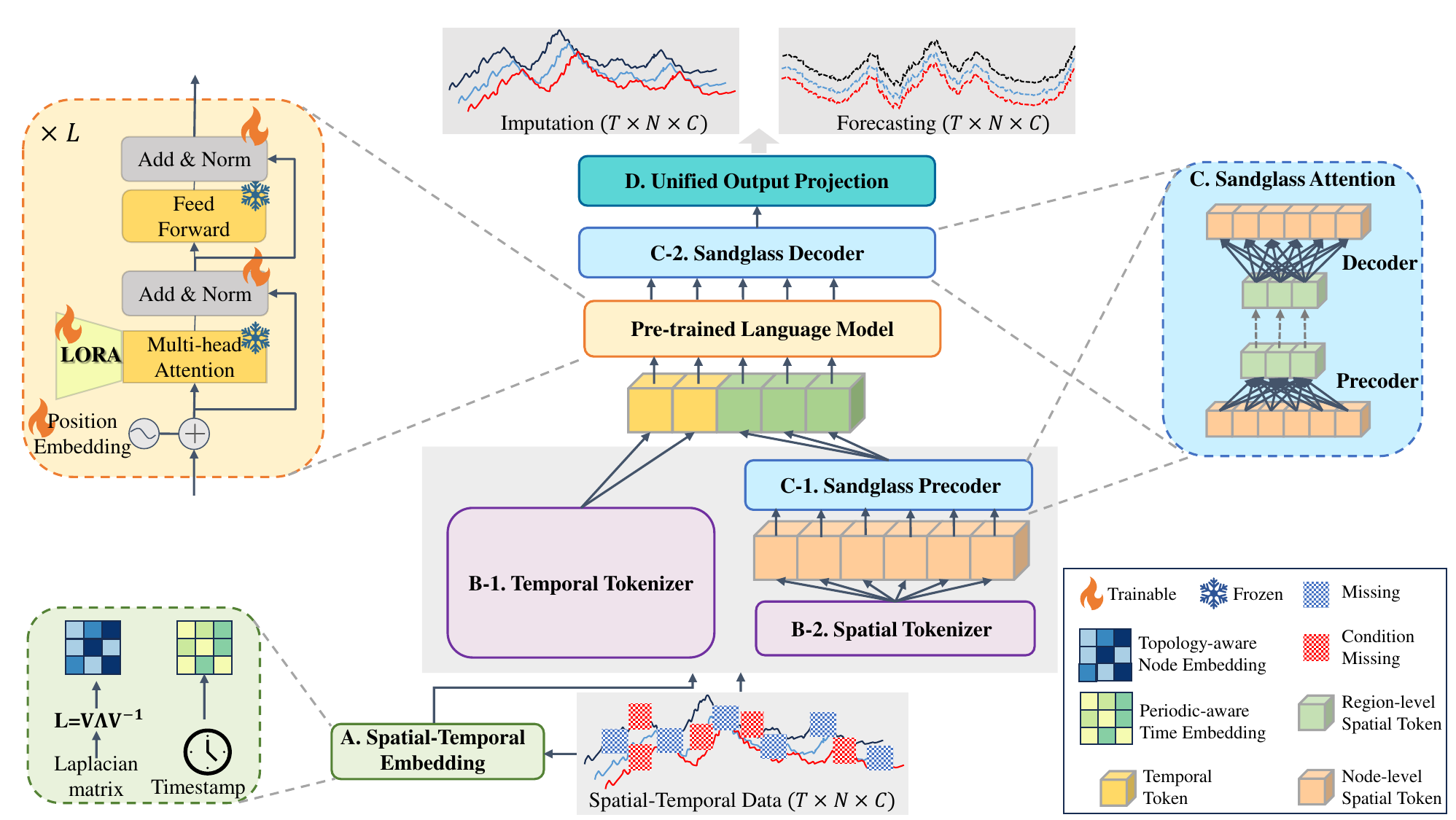} 
  \caption{Model architecture. Module \textbf{A} provides spatial-temporal embeddings for node and temporal information. The spatial-temporal tokenizers \textbf{B} construct temporal and spatial tokens from different perspectives. Module \textbf{C} builds region-level spatial tokens based on the node-level spatial tokens which is the output of \textbf{B-2}. Module \textbf{D} projects the hidden representations to target output. For the PLM, We utilize partial freezing and LoRA to fine-tune the multi-head attention, position embedding and layernorm layers. 
}  
  \label{architecture} 
\end{figure*}

As shown in Figure \ref{architecture}, our model is tailored for accurate spatial-temporal forecasting and imputation by fine-tuning PLM to understand the dependencies and evolving patterns in spatial-temporal data. To achieve this, we first develop a inductive spatial-temporal embedding module to exploit the topology structure and periodicity of data. Based on this, we design temporal and spatial tokenizers to convert the spatial-temporal data into sequential tokens, activating the ability of PLM to understand the inherent spatial, temporal and spatial-temporal correlations of the data represented in the sequential token format. Lastly, we further incorporate sandglass attention module consisting of a precoder and a decoder to not only improve the model efficiency, but also further capture the non-pairwise and higher-order spatial-temporal correlations. 

\subsubsection{Spatial-Temporal Embedding.} 
Spatial-temporal data exhibit heterogeneity at different nodes and time. Thus, we design spatial-temporal embeddings, comprising periodic-aware time embedding and topology-aware node embedding, to represent nodes and time in an inductive manner. 

\noindent \textbf{\textit{Topology-aware node embedding}}: 
 In the design of node embeddings, we need them not only to reflect the static characteristics of each node but also to incorporate the topology structure to express the relationships between nodes. At the same time, the node embeddings must also possess inductive learning capabilities across different graph structures.
 
 Noticing that the graph Laplacian matrix has two properties that align well with our needs, first, the graph Laplacian matrix contains important information about the graph's structure, such as degree and connectivity. Second, the eigenvectors of the Laplacian matrix are orthogonal, which can effectively distinguish different nodes. An initial idea is to use the eigenvectors of the Laplacian matrix as embeddings for each node. However, considering that if the dimensionality of the embeddings is related to the graph's order \(N\), it would not be transferable across different graphs. Therefore, we choose to generate node embeddings based on the eigenvectors corresponding to the \(K\) largest eigenvalues.

Specifically, we define the normalized Laplacian matrix \( \mathbf{L} = \mathbf{I} - \mathbf{D}^{\frac{1}{2}}\mathbf{A}\mathbf{D}^{\frac{1}{2}} \), where \( \mathbf{D}= \sum_{j=1}^{N} \mathbf{A}_{i,j} \)  is the degree matrix and \( \mathbf{I} \) is the identity matrix. The eigendecomposition of \( \mathbf{L} \) yields \( \mathbf{L} = \mathbf{V}\mathbf{\Lambda} \mathbf{V}^{-1} \), with \( \mathbf{V} \) being the matrix of eigenvectors and \( \mathbf{\Lambda} \) the diagonal matrix of eigenvalues. By selecting the eigenvectors corresponding to the top \( K \) largest eigenvalues, we obtain \( \mathbf{V}^{'} \in \mathbb{R}^{N \times K} \). After passing through a linear layer, we broadcast to obtain the topology-aware node embedding \( \mathbf{E}_N \in \mathbb{R}^{T \times N \times d_n} \): 
\begin{align}
 & \mathbf{w}^* = \operatorname{argtop} \limits_{K} (\text{diag}(\mathbf{\Lambda})), \\
 & \mathbf{V}^{'} = \mathbf{V}[:,\mathbf{w}^*],\ \ \ \ \mathbf{E}_N = \mathbf{W}_{ne}\mathbf{V}^{'}+\mathbf{b}_{ne}, 
\end{align}
where "diag" refers to the operation of extracting the diagonal elements of a matrix. \( \mathbf{W}_{ne}\in \mathbb{R}^{K\times d_n} \) and \( \mathbf{b}_{ne}\in \mathbb{R}^{d_n} \) represent the trainable parameters of the linear layers.


\noindent \textit{\textbf{Periodic-aware time embedding}}: 
Spatial-temporal data exhibit periodicity, which is essential for the forecasting and imputation task. To avoid overfitting and underfitting, we select two kinds of periodicity that are moderately coarse including time-of-day and day-of-week to create embedding dictionaries, \( \mathbf{D}_t \in \mathbb{R}^{288 \times d_t} \) and \( \mathbf{D}_w \in \mathbb{R}^{7 \times d_t} \), where \( d_t \) is the dimension of the time embedding.
By looking-up and concatenation along with broadcast operations, we finally obtain the time embedding \( \mathbf{E}_T \in \mathbb{R}^{T \times N \times 2d_t} \). 


\subsubsection{Spatial-Temporal Tokenizer.} 
Utilizing a PLM to process spatial-temporal data, the most crucial step is to transform the spatial-temporal data into tokens that the PLM can handle, as all subsequent processing is based on the PLM's capture of the relationships between these tokens. 
To this end, we meticulously introduce a spatial-temporal tokenizer to generate tokens from both the spatial and temporal dimensions. Specifically,
each node generates a spatial token meanwhile two kinds of temporal tokens are designed to represent the current state and evolving pattern of the spatial-temporal data. Through employing the spatial-temporal tokenizer, the original spatial-temporal graph data is converted into token sequences, enabling the subsequently applied PLM not only to capture the temporal and spatial correlations whithin each kind of tokens but also to capture intertwined spatial-temporal correlations between the spatial and temporal tokens.

\noindent \textit{\textbf{Spatial Tokenizer}}: The spatial tokenizer's objective is to aggregate the temporal information of each node separately, generating a token that represents the state of each node, thereby enabling the PLM to model the complex spatial correlations  between different nodes. For the state of each node, we believe it can be decomposed into a relatively static intrinsic state and a frequently changing dynamic state. The intrinsic state, determined by the node's spatial structure and periodicity, captures the macroscopic characteristics of the node. The dynamic state, derived from historical data, captures the microscopic variations of the node. 

Based on the above analysis, we model the intrinsic states \(\bm{Z}_\text{intrinsic}\in \mathbb{R}^{N\times d_{PLM}}\) using node embeddings and time embeddings, while using historical data to model the dynamic state \(\bm{Z}_\text{dynamic}\in \mathbb{R}^{N\times d_{PLM}}\). To unify the spatial-temporal prediction and imputation tasks, enabling the PLM to detect the patterns of missing is essential. Thus, we further construct mask tokens \(\bm{Z}_\text{mask}\in \mathbb{R}^{N\times d_{PLM}}\) based on the mask matrix \( \bm{\mathcal{M}} \). 
By combining the three elements, we obtain the target node-level spatial tokens \( \bm{Z}_S \in \mathbb{R}^{N\times d_{PLM}}\). Formally, 
\begin{equation}
\label{eq:spatial tokenizer}
\begin{aligned}
 &\bm{Z}_\text{intrinsic} = \text{MLP}([\mathbf{E}_T||\mathbf{E}_N]),\\ &\bm{Z}_\text{dynamic} = \text{MLP}(\mathcal{X}), \\
 &\bm{Z}_\text{mask} = \text{MLP}(\bm{\mathcal{M}}), \\
 &\bm{Z}_S = \text{LayerNorm}(\bm{Z}_\text{dynamic} + \bm{Z}_\text{intrinsic} + \bm{Z}_\text{mask}). \\
\end{aligned}
\end{equation}
In this process, we integrate the temporal dimension into the feature dimension. The MLP consists of two layers of linear with a ReLU activation function in between. \( || \) denotes concatenation along the feature dimension. \( d_{PLM} \) is the token dimension of the PLM.

\noindent \textit{\textbf{Temporal Tokenizer}}: Relying solely on spatial tokens can effectively model the spatial-temporal relationships between nodes, but the model may lack an understanding of the overall state and changing trends of the system. 
Therefore, we design temporal tokenizer that aggregates information from all the nodes for each time step to encapsulate the overall state and changing trends.  
Considering that a single time step cannot capture the sequence's locality and trends, which are essential pieces of information, we draw on the concept from PatchTST ~\cite{patchtst}, merging all time steps into a single patch. Consequently, only one overall state token \(\bm{Z}_\text{state} \in  \mathbb{R}^{1\times d_{PLM}} \) and one overall trend token \(\bm{Z}_\text{trend} \in  \mathbb{R}^{1\times d_{PLM}}\) are produced, rather than generating separate tokens for each time step.

Specifically, we first take the mean of all node to obtain the overall average state \( \bm{\mathcal{\overline X}} \in  \mathbb{R}^{1\times(T \times C)} \). Then, we use the first difference of \( \bm{\mathcal{\overline X}} \), denoted as \( \bm{\mathcal{\overline X}}_\text{trend} \in  \mathbb{R}^{1\times((T-1) \times C)} \), to represent the overall trend. Finally, \( \bm{\mathcal{\overline X}}\) and \(\bm{\mathcal{\overline X}}_\text{trend} \) are respectively combined with the time embeddings. A MLP is further utilized to perform feature transformation, resulting in \(\bm{Z}_\text{state}\)and \(\bm{Z}_\text{trend}\). Then we concatenate these two tokens and apply Layer Normalization to normalize them, thereby obtaining the temporal tokens \(\bm{Z}_{T} \in  \mathbb{R}^{2\times d_{PLM}} \). The above process is defined as follows:
\begin{equation}
\label{temporaltoken}
\begin{aligned}
&\bm{Z}_\text{state} = \text{MLP}(\bm{\mathcal{\overline X}} || \mathbf{E}_T[T-1:,0,:]), \\
&\bm{Z}_\text{trend} = \text{MLP}(\bm{\mathcal{\overline X}}_\text{trend} || \mathbf{E}_T[T-1:,0,:]), \\
& \bm{Z}_T = \text{LayerNorm}([\bm{Z}_\text{state}||\bm{Z}_\text{trend}]), \\
\end{aligned}
\end{equation}
where the \(\mathbf{E}_T[T-1:,0,:] \in  \mathbb{R}^{1\times d_t} \) represents the time embedding of the last time step in the input.


\subsubsection{Sandglass Attention.} 
Taking into account the following two factors, we have developed a sandglass attention module. 1) The number of spatial tokens $N$ is usually large in real-world applications which significantly hampers model efficiency especially the training and inference speed. 2) The spatial tokens primarily encapsulate  information at the node level. However, these node-level spatial tokens alone struggle to capture important non-pairwise and higher-order spatial-temporal correlations. 
Specifically, the proposed sandglass attention (SGA) module first aggregates node-level spatial tokens into fewer region-level spatial tokens thoughr a precoder, enabling the model to capture non-pairwise and higher-order spatial-temporal correlations while enhancing computational efficiency. Then a decoder is employed to restore the original length of spatial tokens. Next, we detail the SGA module. 
 
In the SGA precoder, we aggregate the node-level spatial tokens \(\bm{Z}_S\) into region-level spatial tokens \( \bm{Z}_H \in \mathbb{R}^{M \times d_{LLM}} \) through a learnable query matrix \( \mathbf{H}_{l} \in \mathbb{R}^{M \times d_{PLM}} \):
\begin{equation}
\label{sga1}
 \bm{Z}_H =\text{LayerNorm}( (\text{Attention}(\mathbf{H}_{l}, \bm{Z}_S, \bm{Z}_S) ).
\end{equation}
where $M$ represents the number of region-level spatial tokens, and \(M<N\). Besides, the scaled dot-product-attention is employed in our paper. Formally, 
\begin{equation}
\text{Attention}(\mathbf{Q},\mathbf{K},\mathbf{V}) = \text{Softmax}(\frac{\mathbf{Q}\mathbf{K}^T}{\sqrt{d_k}})\mathbf{V},
\end{equation}

We then feed \( \bm{Z}_H \) into PLM to further capture the spatial-temporal correlations between the region-level spatial tokens and temporal tokens.  Finally, we get the corresponding \( \bm{Z}_H^{'} \in \mathbb{R}^{M \times d_{PLM}} \). To fit with the output size, it is necessary to map the region-level hidden representation back to the node-level by a SGA decoder:  
\begin{equation}
 \bm{Z}_N^{'} = \text{LayerNorm}(( \text{Attention}(\bm{Z}_S, \mathbf{H}_{l}, \bm{Z}^{'}_H)), 
\end{equation}
where \( \bm{Z}_N^{'} \in \mathbb{R}^{N \times d_{PLM}} \) refers to the recovered node-level hidden representation from the PLM.

\subsubsection{Unified Output Projection.} Given that both forecasting and imputation tasks fundamentally involve understanding the complex spatial-temporal dependencies of data, and that the PLM has completed a substantial amount of imputation tasks during pre-train phase, we expect our model, which is designed based on PLM, can handle both of the forecasting and imputation tasks. Thus, we need to enable our model to recognize the presence of missing values. In the Eq. \ref{eq:spatial tokenizer}, we have already enhanced the model's perception of missing values through the mask tokens. Therefore, in the output layer, we unified these two task based on the following projection:
\begin{equation}
\begin{aligned}
 &[\bm{Z}_T^{'},\bm{Z}_H^{'}] = \text{PLM}([\bm{Z}_T || \bm{Z}_H]),\quad \bm{Z}_N^{'} \leftarrow \bm{Z}_H^{'}, \\
 &\mathbf{Y} = {\rm MLP}(\bm{Z}_N^{'} + \bm{Z}_T^{'}[1:,:] + \bm{Z}_S), \\
\end{aligned}
\end{equation}
where \( \mathbf{Y} \in \mathbb{R}^{N \times (T \times C)} \) is the forecasting or imputation result. \(\bm{Z}_T^{'} \in \mathbb{R}^{2 \times d_{PLM}}\) is the hidden representations of temporal tokens. After constructing the tokens and inputting them into the PLM, we obtain the hidden representations. Then, we sum the hidden representation of ecah node with the hidden representation of the overall trend temporal token, and make a residual connection with the node-level spatial tokens to obtain the hidden representation of the target state. Finally, a MLP is used to map the target state hidden representations to our target outputs.
\subsubsection{Model Training.} Given that the PLM has only been exposed to textual data with a significantly different distribution during the pre-training phase and lacks knowledge related to spatial-temporal data, it is necessary to fine-tune the PLM to acquire knowledge of spatial-temporal data and adapt to the distribution patterns of such data. Furthermore, considering that the SGA module needs to learn a query matrix from scratch, this process may lead to overfitting and confusion in spatial correlation if no constraints such as the adjacency matrix are introduced, resulting in a decrease in model performance. Therefore, we designed a constrained loss function to guide and constrain the training of the SGA module. The specific design is detailed in the following text.

\noindent \textit{\textbf{Fine-tuning PLM}}: Based on some existing studies \cite{STLLM,OFA}, we realized that the knowledge of the PLM is primarily stored in the attention modules, while layer normalization and position embeddings are sensitive to the distribution of the data. Therefore, we choose to fine-tune the attention layers to embed spatial-temporal knowledge into the PLM, while also fine-tuning the position embeddings and layer normalization to adapt the PLM to the distribution of spatial-temporal data. To reduce the number of parameters needing updates, we apply Low-Rank Adaptation (LoRA) \cite{lora} to the attention layers. Meanwhile, the position embeddings and layer normalization, with fewer parameters, are fully updated.


\noindent \textit{\textbf{Loss Function}}: We use the L1 loss between the model's output \(Y\) and the ground truth to train our model. Utilizing only this loss function can achieve satisfactory results on the training set. However, to enable the model to make more effective use of the graph structure and to mitigate the phenomenon of overfitting, further design is required. 
Specifically, to ensure that in the SGA module, the learned region-level spatial nodes can reflect the original graph structure \( G \), we design \textit{a structure-aware loss function} \( \mathcal{L}_G \) when aggregating node-level spatial tokens. Furthermore, to prevent overfitting that might cause some node information to be ignored (attention weights are close to zero), we introduce \textit{a regularization term} \(\mathcal{L}_R \). The sum of these two components forms constraint loss function \( \mathcal{L}_C = \mathcal{L}_G +  \mathcal{L}_R \). Assuming \( S \in \mathbb{R}^{M \times N} \) represents the attention weights between each region-level spatial token and node-level spatial token, \(\mathcal{L}_G \) and \(\mathcal{L}_R \) can be expressed as:
\begin{equation}
\setlength\abovedisplayskip{0pt}
\setlength\belowdisplayskip{3pt}
\label{constraintloss}
\begin{aligned}
 &\mathcal{L}_G = -\sum_{m=1}^M \sum_{i,j(i\neq j)} {S}_{m,i} {S}_{m,j} A_{i,j}, \\
 &\mathcal{L}_R = -\log \pi(\text{Softmax}(\sum_{m=1}^M S_{m,:}) | \bm{\alpha}),
\end{aligned}
\end{equation}
where \( \pi \) is the Dirichlet distribution. \( {\bm{\alpha}} \in \mathbb{R}^{N} \) is the parameter of the Dirichlet distribution. \( S_{m,:} \in \mathbb{R}^{N} \) denotes the vector of attention weights for the \( m \)-th region token across all node tokens, which satisfies the properties: \(\sum_{i=1}^N S_{m,i} = 1 \). 

It is clear that the more edges there are, the smaller the minimum value of \(L_G\) can be. Consider a complete graph of size P, where:
\begin{align}
\setlength\abovedisplayskip{0pt}
\setlength\belowdisplayskip{0pt}
\mathcal{L}_G &=  -\sum_{m=1}^M \sum_{i,j(i\neq j)} {S}_{m,i} {S}_{m,j}\cdot 1, \\
\notag
    &= -\sum_{m=1}^M \sum_{i} {S}_{m,i}(1-{S}_{m,i})= -M + \Vert {S}\odot {S} \Vert_1.
\end{align}
The \(\odot\) denotes the Hadamard Product. It is evident that minimizing the objective function \(\mathcal{L}_G\) is achieved by evenly distributing the values of \(S\), such that \(S_{m,i} = \frac{1}{P}\). Under this condition, \(\mathcal{L}_G\) simplifies to \(-\frac{M(P-1)}{P}\). Shifting the discussion to a regular graph of order \(N\), a lower \(\mathcal{L}_G\) facilitates the SGA module's ability to concentrate on larger complete subgraphs or those with robust connectivity, effectively integrating the graph's structural properties into the model. 

The essence of \(\mathcal{L}_R\) is simple: the expected value of \(\pi\) is \( \bm{\alpha}/{\sum_i \bm{\alpha}_i} \), indicating that a larger \(-\mathcal{L}_R\) drives \(\text{Softmax}(\sum_{m=1}^M S_{m,:})\) closer to this expected value. By tuning \(\bm{\alpha}\), we indirectly control the attention weights distribution. Setting \(\bm{\alpha} = \{1.05\}^N\) (a small value greater than 1) ensures equal aggregation of nodes. Yet, to prioritize significant nodes, we refine \(\bm{\alpha}\) by adding the node degrees: \(\bm{\alpha} = \{1.05\}^N + \text{Softmax}(\text{diag}(D))\).

\section{Experiments}
\label{section:experiments}
\subsection{Datasets}
The experiments conducted on four traffic datasets (PEMS03, PEMS04, PEMS07, PEMS08) \cite{chen2001freeway}.

To assess the model's imputation capability, we generated two types of missing data patterns, RM (random missing) and CM (spatial-temporal continuity missing), on the PEMS08 dataset, each with a 70\% missing rate. Additionally, following the self-supervised training approach of CSDI ~\cite{CSDI}, we generated condition missing for training. Condition missing refers to the artificially created missing data, used to generate training samples.

The details of the four datasets and missing pattern are provided in supplementary material.

\begin{table*}[h]
		\setlength{\abovecaptionskip}{0cm} 
		\setlength{\belowcaptionskip}{-0.2cm}
  \setlength{\tabcolsep}{3pt}
  \caption{Forecasting performance. There are two kinds of baselines, including six conventional deep learning models and three PLM-based models.}
  \label{forecasting}
  \centering
\resizebox{\linewidth}{!}{
  \begin{tabular}{ccccccccccccc}
    \toprule
     \multirow{2}{*}{\textbf{Model}} & \multicolumn{3}{c}{\textbf{PEMS03}}  & \multicolumn{3}{c}{\textbf{PEMS04}} & \multicolumn{3}{c}{\textbf{PEMS07}} & \multicolumn{3}{c}{\textbf{PEMS08}}  \\
     \cmidrule(r){2-4} \cmidrule(r){5-7}  \cmidrule(r){8-10} \cmidrule(r){11-13}  
    & \textbf{MAE} & \textbf{RMSE} & \textbf{MAPE} & \textbf{MAE} & \textbf{RMSE} &\textbf{MAPE} & \textbf{MAE} & \textbf{RMSE} & \textbf{MAPE} & \textbf{MAE} & \textbf{RMSE} &\textbf{MAPE} \\
    \midrule
    LSTM    &20.62    &33.54    &28.94\%    &26.81    &40.74    &22.33\%    &29.71    &45.32    &14.14\%     &22.19     &33.59     &18.74\%\\ 
    ASTGCN    &17.85    &29.88    &17.65\%    &22.42    &34.75    &15.87\%    &25.98    &39.65    &11.84\%     &18.86     &28.55     &12.50\%\\ 
    AGCRN    &15.98    &28.25    &\underline{15.23\%}    &19.83    &32.26    &12.97\%    &22.37    &36.55    &9.12\%     &15.95     &25.22     &10.09\%\\ 
    SSTBAN   &15.90    &26.11    &17.27\%   &18.88    &31.10    &12.56\%    &20.17    &33.45   &8.93\%     &14.39     &24.19     &10.06\%\\ 
    PDFormer &\underline{14.74}    &25.59    &15.35\%    &\underline{18.31}    &\textbf{29.97}    &\underline{12.10\%}    &\underline{19.83}    &\underline{32.87}    &\underline{8.53\%}    &\underline{13.58}  &23.51     &\underline{9.05\%} \\ 
    iTransformer     &19.48 &31.20 	&17.84\%    &22.56  &35.21 &16.29\%    &24.70 	&37.91 	&11.40\%     &20.05 	&31.90 	&11.99\% \\   \midrule
    OFA      &20.96    &33.43    &19.11\%      &27.37    &42.99    &17.97\%    &30.53    &47.51    &12.98\%    &21.89 	  &34.63 	 &13.30\%  \\ 
    STGLLM    &15.26    &\textbf{24.11}    &15.73\%    &20.00    &32.11    &13.69\%    &21.98    &35.02    &9.72\%     &15.53     &24.74     &10.15\% \\ 
    STLLM     &17.25    &27.25    &22.96\%    &19.00    &30.35    &13.55\%    &21.48    &34.07    &10.20\%     &14.67     &\underline{23.50}     &10.63\% \\ 
    \textbf{STD-PLM}  &\textbf{14.59} 	&\underline{25.36} 	&\textbf{14.92\%}  &\textbf{18.16} 	&\underline{30.21} 	&\textbf{11.89\%}   &\textbf{19.25} 	&\textbf{32.84} 	&\textbf{8.06\%}     &\textbf{13.31} 	&\textbf{23.19} 	&\textbf{8.84\%}\\ 
     \bottomrule
  \end{tabular}
}
\end{table*}

\subsection{Experimental Settings}

We follow the approach of ASTGCN \cite{ASTGCN} by partitioning the dataset into training set, validation set, and testing set with a ratio of 6: 2: 2, and employ a sliding window of size 12 to construct samples that forecast the next 12 steps based on the historical 12 steps.

To comprehensively compare the performance of models, we selected nine representative baselines for the forecasting task, including Long Short-Term Memory network(LSTM) \cite{lstm}, ASTGCN \cite{ASTGCN}, AGCRN \cite{AGCRN}, SSTBAN \cite{SSTBAN}, PDFormer \cite{PDFormer}, iTransformer \cite{iTransformer}, OFA \cite{OFA}, STLLM \cite{STLLM} and STGLLM \cite{STGLLM}. For the imputation task, we also chose four well-known baselines, which are Brits \cite{BRITS}, E2GAN \cite{E2GAN}, mTAN \cite{mTAN}, and PriSTI \cite{PRISTI}.

The experiments were conducted on NVIDIA A40 GPUs with a PyTorch version of 2.1.2. Using an AdaW optimizer with a learning rate of \( 1 \times 10^{-3} \). The training epoch is 500 with a 50 epochs early stopping mechanism. We use GPT-2 \cite{gpt2} as the PLM and utilized only its initial three layers. STD-PLM employs the hyperparameters identified through tuning on the PEMS08 validation set for all datasets. The detail of experiments settings are depicted in supplementary material.

\begin{table}[h]
		\setlength{\abovecaptionskip}{0cm} 
		\setlength{\belowcaptionskip}{-0.2cm}
  \setlength{\tabcolsep}{3pt}
  \caption{Imputation performance on PEMS08.}
  \label{imputation}
  \centering
  \begin{tabular}{ccccccccccccc}
    \toprule
     \multirow{2}{*}{\textbf{Model}}   & \multicolumn{3}{c}{\textbf{RM 70\%}}  & \multicolumn{3}{c}{\textbf{CM 70\%}}  \\
     \cmidrule(r){2-4} \cmidrule(r){5-7} 
    & \textbf{MAE} & \textbf{RMSE} & \textbf{MAPE} & \textbf{MAE} & \textbf{RMSE} &\textbf{MAPE}  \\
    \midrule
    BRITS    &29.56      &41.52      &\underline{10.79\%}    &38.33 	&53.60 	&\underline{13.90\%}  \\ 
    E2GAN    &27.55 	    &41.99 	    &17.52\%       &31.58 	&50.24 	&19.17\% \\ 
    mTAN     &\underline{21.23} 	&\underline{33.98} 	&12.89\%      &31.42 	&49.10 	&18.96\% \\ 
    PriSTI   &25.54	    &36.00	        &16.73\%    &\underline{22.93}	&\underline{40.90}	&14.43\%  \\ 
    \textbf{STD-PLM}   &\textbf{14.36} 	&\textbf{23.20} 	&\textbf{9.58\%}       &\textbf{22.69} &\textbf{39.66} 	&\textbf{13.82\%}\\  
     \bottomrule
  \end{tabular}
\end{table}

\subsection{Overall Performance}
We measure the model's performance using three widely used metrics for regression tasks: Mean Absolute Error (MAE), Root Mean Square Error (RMSE), and Mean Absolute Percentage Error (MAPE). Table \ref{forecasting} shows the forecasting performances comparison between our proposed STD-PLM and baselines. Bold indicates the best results, while underline denotes the second-best results. From Table \ref{forecasting}, we can observe that: 1) The accuracy of the time series models such as LSTM, iTransformer and OFA is not good enough compared to methods that can handle spatial-temporal correlations in data. This indicates that relying solely on temporal correlation is insufficient to achieve high-precision forecasting; 2) The results of PLM-based models OFA, STGLLM, and STLLM demonstrate the necessity of our design, as simply using the PLM to process spatial-temporal data does not lead to optimal performance. 3) STD-PLM achieve the best or second-best results on the all datasets.

Table \ref{imputation} presents the imputation performances comparison between our proposed STD-PLM and baselines. Based on the comparison, we can observe that: 1)Our method achieves state-of-the-art performance. 2) STD-PLM exhibits a substantial improvement over other baseline models in the RM 70\% case. This implies that STD-PLM can fully harness the imputation capabilities of the PLM for spatial-temporal data. 

In summary, the experimental results demonstrate that STD-PLM has strong performance and can achieve high accuracy in both forecasting and imputation tasks.

\subsection{Few-shot and Zero-shot Performance}
\label{Few-shot and Zero-shot Performance}
To extensively evaluate the performance of PLM on spatial-temporal tasks, we conduct few-shot and zero-shot experiments. The few-shot experiments evaluate the forecasting performance of STD-PLM on the PEMS04 and PEMS08 datasets using only the first 5\%, 10\%, and 20\% of the training samples, with results summarized in Table \ref{fewshot}. The zero-shot experiments evaluate the forecasting  performance of the trained model when applied to other datasets, with the results compiled in Table \ref{zeroshot}.

From Table \ref{fewshot}, it can be observed that we only require 5\% of the training samples to achieve performance comparable to that of LSTM trained on the full dataset. When the training samples are increased to 20\%, the performance surpasses that of ASTGCN. This indicates that STD-PLM has excellent few-shot learning capabilities, which supports its application in scenarios where data is scarce. The results in Table \ref{zeroshot} demonstrate that STD-PLM can exhibit acceptable performance when directly transferred to datasets with different temporal scopes and graph structures from the training set, without undergoing any training. 

\begin{table*}[h]
    \setlength{\abovecaptionskip}{0cm} 
    \setlength{\belowcaptionskip}{-0.2cm}
    \setlength{\tabcolsep}{2pt}
    \centering
    \begin{floatrow}
        \resizebox{\linewidth}{!}{
            \capbtabbox{
                  \begin{tabular}{ccccccc}
                    \toprule
                     \multirow{2}{*}{\textbf{Ratio}} & \multicolumn{3}{c}{\textbf{PEMS04}}  & \multicolumn{3}{c}{\textbf{PEMS08}}   \\
                     \cmidrule(r){2-4} \cmidrule(r){5-7}  
                    & MAE & RMSE & MAPE & MAE & RMSE &MAPE \\
                    \midrule
                    5\% &27.73  &42.12 	&20.62\% &22.56 &34.35 &16.91\% \\ 
                    10\% &25.11 &38.68 	&17.65\% &19.72 	&31.35 	&13.27\%   \\ 
                    20\% &21.16 	&34.03 	&13.92\%  &16.58 	&27.03 	&10.78\%     \\ 
                     \bottomrule
                  \end{tabular}
                }{
                  \caption{Few-shot performance.}
                  \label{fewshot}
                }
                
            \capbtabbox{
                  \begin{tabular}{cccc}
                    \toprule
                    & \textbf{MAE} & \textbf{RMSE} & \textbf{MAPE}  \\
                    \midrule
                    PEMS04\(\xrightarrow{}\)PEMS08 &29.52 	&45.63 	&22.92\% \\ 
                    PEMS08\(\xrightarrow{}\)PEMS04  &25.32 	&37.80 	&25.32\% \\ 
                    PEMS07\(\xrightarrow{}\)PEMS03  &23.72 	&36.94 	&41.59\%  \\ 
                    PEMS03\(\xrightarrow{}\)PEMS07  &34.72 	&52.66 	&20.31\%  \\ 
                     \bottomrule
                  \end{tabular}
                }{
                      \caption{Zero-shot performance.}
                      \label{zeroshot}
                }
        }
    \end{floatrow}
\end{table*}





\begin{table*}[h]
    \setlength{\abovecaptionskip}{0cm} 
    \setlength{\belowcaptionskip}{-0.2cm}
    \setlength{\tabcolsep}{2pt}
    \centering
    \begin{floatrow}
        \resizebox{\linewidth}{!}{
            \capbtabbox{
                  \begin{tabular}{ccccccccccccc}
                    \toprule
                     \multirow{2}{*}{\textbf{Method}}   & \multicolumn{3}{c}{\textbf{PEMS08}}  & \multicolumn{3}{c}{\textbf{PEMS08 CM 70\%}}  \\
                     \cmidrule(r){2-4} \cmidrule(r){5-7}   
                     & MAE & RMSE &MAPE & MAE & RMSE &MAPE \\
                    \midrule
                    w/o TT  &13.81 	&23.52 	&9.03\%  &22.82 	&\textbf{39.50} 	&14.35\% \\ 
                    w/o CLoss    &13.53 	&23.69 	&8.97\%   &23.62 	&40.05 	&14.65\%  \\ 
                    w/o PLM   &13.63 &23.39 &9.15\% &24.05 	&41.15 	&15.34\% \\ 
                    \textbf{STD-PLM}     &\textbf{13.31}	&\textbf{23.19} &\textbf{8.84\%} &\textbf{22.69} &\underline{39.66} 	&\textbf{13.82\%} \\  
                     \bottomrule
                  \end{tabular}
                }{
                      \caption{Ablation study.}
                      \label{ablation}
                }
                
            \capbtabbox{
                      \begin{tabular}{ccccccccccc}
                        \toprule
                         \multirow{2}{*}{\textbf{Method}}   & \multicolumn{2}{c}{\textbf{PEMS03}}  & \multicolumn{2}{c}{\textbf{PEMS07}}  \\
                         \cmidrule(r){2-3} \cmidrule(r){4-5}   
                         & Time & Memory & Time & Memory  \\
                        \midrule
                        w  SGA  &7.40 &8554  &9.15 &15020  \\ 
                        w/o SGA &17.96  &15366   &52.82 &29718   \\  
                         \bottomrule
                      \end{tabular}
                }{
                      \caption{inference consumption.}
                      \label{SGA cost}
                }
        }
    \end{floatrow}
\end{table*}



\subsection{Ablation Study}

We implement ablation study to evaluate the effectiveness of each component. The ablation experiments compared our proposed model with the following variants: 1) w/o TT, removing the temporal tokens;  2) w/o CLoss, removing the \(\mathcal{L}_C\) in Eq.\ref{constraintloss}; 3) w/o PLM, replacing the PLM with a transformer encoder of the same layer and hidden dimension.

We summerize the results of the ablation study in Table \ref{ablation}. Based on the results, we can identify the following: 1) Temporal Token and \(\mathcal{L}_C\) can enhance the model's forecasting and imputation performance with a relatively small computational cost. 2) Fine-tuning the PLM can achieve better results than a transformer encoder with the same parameter scale that updates all parameters during training. That means that the knowledge embedded within the PLM is useful for spatial-temporal tasks. 

Additionally, we test the impact of the SGA on efficiency. The results of the experiments are summarized in Table \ref{SGA cost}, we record the time (s) and gpu memory (MiB) usage to infer the entire test set. From this, we can observe that the efficiency improvement brought by the SGA is substantial.

\section{Conclusion}
\label{section:conclusion}
In this paper, we propose STD-PLM, a unified framework for spatial-temporal forecasting and imputation based on PLM. Through our explicitly designed spatial-temporal tokenizers and spatial-temporal embeddings, STD-PLM can effectively understand both spatial and temporal properties of spatial-temporal data. Additionally, we introduce a SGA module to significantly reduces computational costs by constructing region-level spatial tokens. Extensive experiments demonstrate that STD-PLM exhibits competitive performance. Our work suggests that constructing a unified pre-trained spatial-temporal model based on PLM is promising.

\newpage
\bibliography{main}

\newpage
\appendix

\section{Limitations}
\label{appendix:limitations}
Due to computational limitations, this paper did not further test whether PLM could yield better results. In theory, the recently emerged PLM with 7 billion, 30 billion, or even more parameters should possess stronger capabilities for processing spatial-temporal data. Additionally, although this paper proposed a unified pre-training framework, we lack sufficient data to test whether the model, after being trained on a large-scale spatial-temporal dataset, would exhibit emergent phenomena similar to those of PLM, leading to a substantial improvement in performance. These are some of the limitations of our research presented in this paper, and we hope that future work will be supported by adequate computational power and data to further explore these possibilities.

\section{Broader Impacts}
\label{appendix:broader impacts}
Our proposed model can demonstrate relatively accurate spatial-temporal forecasting and imputation capabilities with only a small number of training samples or even without samples, which implies that our model can support a broader range of deployment and applications compared to other models. Moreover, the fast inference speed indicates that our model can sustain real-time interactive applications. Based on this, our work can provide rapid-response spatial-temporal forecasting services to a wider area, such as offering traffic dispatch decision support and travel route planning through accurate and swift traffic flow forecasting for different regions, thereby enhancing the efficiency of societal operations. However, a potential negative impact of this work is that inaccurate forecasting could lead to erroneous decision-making, resulting in adverse social consequences.

\section{Experimental Details}
\label{appendix:EXPERIMENTAL DETAILS}
\subsection{Baselines}
To compare model performance, we selected 7 forecasting models and 4 imputation models as baselines. Detailed introductions to the baselines are as follows:
\begin{itemize}
    \item LSTM: A special type of RNN (Recurrent Neural Network) that mitigates the vanishing and exploding gradient problems by incorporating memory cells and forget gates, enabling the processing of long-term dependencies in sequences.
    \item ASTGCN: A traffic forecasting model based on attention and convolutional networks. It captures the dynamic correlations of data through spatial-temporal attention and combines spatial-temporal graph convolutions to capture temporal features and spatial features.
    \item AGCRN: An adaptive graph convolutional neural network for traffic forecasting. It optimizes the conventional GCN (Graph Convolutional Network) by proposing a dynamic graph convolution algorithm that learns adaptively from the data. It learns transformation matrices for each node separately to better handle different spatial-temporal patterns of nodes. By integrating the dynamic graph convolution with GRU (Gated Recurrent Unit), it efficiently processes spatial-temporal data.
    \item SSTBAN: An attention based long-term spatial-temporal prediction model using an autoencoder architecture. It designs a bottleneck attention that can effectively reduce the time complexity of the model's attention process through some reference points. At the same time, the model also combines a mask autoencoder to effectively enhance the encoding ability of the model for spatial-temporal data through completion tasks.
    \item PDFormer: An attention-based traffic forecasting model. It fully considers the shortcomings of conventional GNN (Graph Neural Network) models in handling dynamic spatial-temporal dependencies, long-range dependencies, and information propagation delays in traffic systems. By combining geographical distance and traffic similarity to generate semantic and geographical neighbors, it enables spatial-temporal attention to handle both short-range and long-range dynamic spatial-temporal dependencies. Additionally, it uses the k-shape clustering algorithm to calculate similar historical traffic patterns for each sample, thus simulating information propagation delays.
    \item iTransformer: A transformative approach to time series forecasting that inverts the traditional Transformer model. By focusing on variate-centric embeddings and self-attention mechanisms, it adeptly captures multivariate correlations and learns robust nonlinear representations, achieving state-of-the-art results with enhanced efficiency and generalization.
    \item OFA: A time series model based on PLM. It processes time series data by segmenting it into patches and converting it into tokens for PLM processing. It fine-tunes only the layer normalization and position encoding of PLM to retain its general capabilities.
    \item STGLLM: A spatial-temporal forecasting model based on PLM. It enhances PLM's understanding of spatial-temporal data using time embedding and other external information, as well as prompts. It also fine-tunes only the layer normalization and position encoding of PLM.
    \item STLLM: A spatial-temporal prediction model based on PLM employs a special partial fine-tuning technique for the PLM. It fine-tunes only the Layer Normalization for the initial layers of the PLM, while for the subsequent layers, both the attention layers and Layer Normalization are fine-tuned simultaneously.
    \item BRITS: A multivariate time series imputation model based on bidirectional RNNs (Recurrent Neural Networks). It employs a delayed error design for self-supervised training of the model. By using the consistency constraints of bidirectional RNNs, it addresses the error accumulation issue of unidirectional RNNs. It integrates both history-based and feature-based estimations.
    \item E2GAN: A multivariate time series imputation model based on GAN (Generative Adversarial Network). The generator uses an autoencoder structure, mapping incomplete inputs to a latent space with added random noise, and then gradually decoding them into complete sequences. Both the encoding and decoding modules of the generator and the discriminator use GRUI (Gated Recurrent Unit for data Imputation), a GRU that incorporates a time decay factor.
    \item mTAN: A multivariate time series imputation model based on VAE (Variational Autoencoder) and attention. It maps irregularly sampled sequences into a latent space using attention. It samples latent states from the latent space and then uses attention and RNN to obtain the complete sequence.
    \item PriSTI: A spatial-temporal data imputation model based on Diffusion Model and attention. It transforms the imputation problem into a generation problem by filling missing parts with random noise and then using the denoising module of the diffusion model to convert the noise into real values. It designs a conditional feature extraction module to provide a coarse-grained spatial-temporal dependency context prior for the spatial-temporal attention mechanism in the denoising module, thus enabling more accurate imputation.
\end{itemize}

\subsection{Datasets}
\label{appendix:dataset}
The detail of four dataset used in our experiments are summarized in Table \ref{tab:dataset}. Data is sampled every 30 seconds and compiled into 5-minute intervals. The table reveals that the datasets contain a small number of edges, indicating that all four have comparatively sparse graph structures.

\begin{table*}[h!]
    \caption{Datasets}
    \label{tab:dataset}
    \centering
    \begin{tabular}{cccccc}
    \toprule
        \textbf{Datasets} & \textbf{Samples} & \textbf{Nodes} & \textbf{Edges} & \textbf{Interval} & \textbf{Time Range}\\
     \midrule
        PEMS03 & 26208 & 358 & 547 & 5min & 09/01/2018-11/30/2018\\
        PEMS04 & 16992 & 307 & 340 & 5min & 01/01/2018-02/28/2018\\
        PEMS07 & 28224 & 883 & 866 & 5min & 05/01/2017-08/31/2017\\
        PEMS08 & 17856 & 170 & 295 & 5min & 07/01/2016-08/31/2016\\ \midrule
    \bottomrule
    \end{tabular}

\end{table*}

\subsection{Missing Patterns}
\label{appendix:missingpatterns}
We construct missing data on the complete dataset to train and test the model's imputation capabilities. The experiment is designed with two different types of missing data patterns, namely random missing(RM) and spatial-temporal continuous missing(CM), which are defined as follows:
\textbf{RM:} A missing pattern that is unrelated to both time and space, and is entirely random. In constructing this pattern, it is only necessary to randomly mask a portion of the data according to the missing rate.
\textbf{CM:} A pattern of missingness that has a strong correlation with both time and space, characterized by simultaneous missing data in a specific area over a certain period. We define three consecutive time steps as one patch and use graph clustering algorithms to divide the graph into several smaller regions. When constructing the missing data, we randomly select patches and small regions, masking all the data within them.
\subsection{Model Configurations}
The hyperparameters used in our model for all experiments are the same, as summarized in Table \ref{MODEL CONFIGURATIONS}:

\begin{table*}[h!]
  \caption{MODEL CONFIGURATIONS.}
  \label{MODEL CONFIGURATIONS}
  \centering
\resizebox{\linewidth}{!}{
  \begin{tabular}{cccc}
    \toprule
     \textbf{Module}  & \textbf{Hyper Paremeters} & \textbf{Note} & \textbf{Value} \\
     \midrule 
    \multirow{3}{*}{Spatial-Temporal  embedding} & \(d_t\) & the dimension of time embedding & 64 \\ 
                                                         & \(d_n\) & the dimension of topology-aware node embedding & 64 \\
                                                         & \(K\)   & the number of truncated eigenvectors & 64 \\ \midrule
     \multirow{2}{*}{SGA}       & \(M\) & the number of region-level tokens & 128 \\ 
                                                         & \(d_H\) & the dimension of hidden space in SGA & 128 \\ \midrule        
      \multirow{2}{*}{PLM}                     & \(d_{PLM}\)        & the dimension of the word embeddings in  PLM & 768 \\ 
                                                         & Layers & the number of PLM layers utilized  & 3 \\ \midrule  
     \bottomrule
  \end{tabular}
}
\end{table*}

\subsection{Parameter Efficacy} 
\label{appendix:Parameter Efficacy}

An important question to consider is whether the PLM significantly increases the computation cost. We have summarized the computation cost of main baselines and STD-PLM in Table \ref{computation cost}. It can be observed from the table that, although our model has a large number of parameters, only 2.69\% of them require training, and our model also has the fastest inference speed. The incorporation of PLM does indeed substantially increase the number of parameters, which leads to higher GPU memory usage. However, thanks to the relatively simple structure of PLM, all operations can be highly parallelized, thus not significantly impacting the training and inference time. Additionally, the regional-level tokens constructed by our SGA module also reduce the number of tokens, thereby enhancing computation speed.
\begin{table*}[h!]
  \caption{The computation cost on the PEMS08 with the batchsize=64.}
  \label{computation cost}
  \centering
\resizebox{\linewidth}{!}{
  \begin{tabular}{ccccc}
    \toprule
     \textbf{Method}  & \textbf{Parameters} & \textbf{Trainable Parameters} & \textbf{Trainable Ratio(\%)}  & \textbf{Interface Time(s/epoch)}  \\ \midrule
    PriSTI   &937646 &931246 &99.31 & 2193.27   \\ 
    PDFormer &537565 &531165 &98.80 &9.49     \\ 
    STGLLM   &64014315 &1016808 &1.58  &8.36 \\ 
    STLLM   &82640908 &42494476 &51.42  &24.42 \\ 
    \textbf{STD-PLM (ours)}   &66136848 &3128461 &4.73  &4.38  \\ \midrule
     \bottomrule
  \end{tabular}
}
\end{table*}

\section{Error Bars}
\label{appendix:Error Bars}
All experiments were conducted three times, and the final metrics were taken as the average of the three trials. We present the standard deviations for STD-PLM in the forecasting and imputation experiments in Table \ref{Standard deviations} and Table \ref{Standard deviations imputation}.

\begin{table*}[h!]
    \setlength{\abovecaptionskip}{0cm} 
    \setlength{\belowcaptionskip}{-0.2cm}
    \setlength{\tabcolsep}{2pt}
    \centering
    \begin{floatrow}
        \resizebox{\linewidth}{!}{
            \capbtabbox{
                  \begin{tabular}{ccccc}
                    \toprule
                    \textbf{Metric}  & \textbf{PEMS03} & \textbf{PEMS04} & \textbf{PEMS07}  & \textbf{PEMS08}  \\ \midrule
                    MAE        &\(14.59\pm 0.0163\)   &\(18.16\pm 0.0307\)     &\(19.25\pm 0.1018\)    &\(13.31\pm 0.0249\)   \\ 
                    RMSE       &\(25.36\pm 0.1582\)   &\(30.21\pm 0.0193\)     &\(32.84\pm 0.0815\)    &\(23.19\pm 0.0857\)     \\ 
                    MAPE(\%)   &\(14.92\pm 0.0727\)   &\(11.89\pm 0.0429\)     &\(8.06\pm 0.0383\)    &\(8.84\pm 0.0214\) \\ \midrule
                     \bottomrule
                  \end{tabular}
             }{
               \caption{Standard deviations of STD-PLM in the forecasting experiment.}
                \label{Standard deviations}
             }
            \capbtabbox{
                  \begin{tabular}{ccc}
                    \toprule
                    \textbf{Metric}   & \textbf{RM 70\%}   & \textbf{CM 70\% } \\ \midrule
                    MAE           &\(14.36\pm 0.0661\)        &\(22.69\pm 0.0768\)   \\ 
                    RMSE         &\(23.20\pm 0.2201\)        &\(39.66\pm 0.5531\)     \\ 
                    MAPE(\%)      &\(9.58\pm 0.0846\)         &\(13.82\pm 0.1213\) \\ \midrule
                     \bottomrule
                  \end{tabular}
             }{
                  \caption{Standard deviations of STD-PLM in the imputation experiment on PEMS08.}
                  \label{Standard deviations imputation}
             }
        }
\end{floatrow}
\end{table*}





\end{document}